\documentclass[11pt,a4paper]{article}
\usepackage{times,latexsym}
\usepackage{url}
\usepackage[T1]{fontenc}
\usepackage{amsmath}
\usepackage{amsthm}
\usepackage{amsfonts}
\usepackage{algorithm}
\usepackage{graphicx}
\usepackage{subcaption}
\usepackage{bm}
\usepackage{comment}
\usepackage{multirow}
\usepackage[dvipsnames]{xcolor}

\newcommand{\red}[1]{\textcolor{red}{{#1}}}
\newcommand{\blue}[1]{\textcolor{blue}{{#1}}}
\newcommand{\green}[1]{\textcolor{ForestGreen}{{#1}}}
\usepackage{soul}

\usepackage[acceptedWithA]{tacl2018v2}

\usepackage{xspace,mfirstuc,tabulary}

\newif\iftaclinstructions
\taclinstructionsfalse 
\iftaclinstructions
\renewcommand{\confidential}{}
\renewcommand{\anonsubtext}{(No author info supplied here, for consistency with
TACL-submission anonymization requirements)}
\newcommand{\instr}
\fi

\iftaclpubformat 

\else

\fi

\title{Controllable Summarization with Constrained Markov Decision Process}

\author{Hou Pong Chan\textsuperscript{\rm 1}, 
Lu Wang\textsuperscript{\rm 2}, {\rm and}
Irwin King\textsuperscript{\rm 3} \\
\textsuperscript{\rm 1}University of Macau, Macau SAR, China\\
\textsuperscript{\rm 2}University of Michigan, Ann Arbor, MI, USA\\
\textsuperscript{\rm 3}The Chinese University of Hong Kong, Hong Kong SAR, China\\
{\sf \textsuperscript{\rm 1}hpchan@um.edu.mo}\\
{\sf \textsuperscript{\rm 2}wangluxy@umich.edu}\\
{\sf \textsuperscript{\rm 3}king@cse.cuhk.edu.hk}\\
}

\date{}

\begin{document}
\maketitle
\begin{abstract}
We study \emph{controllable text summarization} which allows users to gain control on a particular attribute (e.g., length limit) of the generated summaries. In this work, we propose a novel training framework based on \emph{Constrained Markov Decision Process} (CMDP), which conveniently includes a {\it reward function} along with a set of {\it constraints}, to facilitate better summarization control. 
The reward function encourages the generation to resemble the human-written reference, while the constraints are used to explicitly prevent the generated summaries from violating user-imposed requirements. 
Our framework can be applied to control important attributes of summarization, including \emph{length}, \emph{covered entities}, and \emph{abstractiveness}, as we devise specific constraints for each of these aspects. 
Extensive experiments on popular benchmarks show that our CMDP framework helps generate informative summaries while complying with a given attribute's requirement\footnote{Our source code is available at \url{https://github.com/kenchan0226/control-sum-cmdp}}. 
\end{abstract}

\section{Introduction}
Text summarization aims to condense the information of an input document into a concise summary.
Although recently neural abstractive summarization models have achieved promising performance~\cite{DBLP:conf/acl/PtrGen17, DBLP:conf/iclr/PaulusXS18}, they do not allow users to indicate their preference to control different aspects of the generated summaries. 
Controllable summarization has many use cases.
For instance, it can summarize product descriptions to fit within a word limit in online advertising.
In another example, teachers can demonstrate the technique of paraphrasing important information by showing a system-generated summary with high abstractiveness. 
Controllable summarization can also complement information retrieval systems, for example, to only generate summaries covering the entities that users are interested in. Figure~\ref{figure:intro-example} illustrates one such usage, where our proposed model produces distinct abstractive summaries of the same source document, focusing on different input entities. 

\begin{figure}[t]
\small
\centering
\fontsize{9}{10}\selectfont
\begin{tabular}{|p{0.92\columnwidth}|}
\hline
\textbf{Document: } \\ \hline
Marseille, France (CNN) The French prosecutor leading an investigation into \blue{the crash of Germanwings Flight 9525} insisted Wednesday that he was not aware of any video footage from on board the plane. \red{Marseille prosecutor Brice Robin, in charge of the criminal inquiry into the crash, told CNN that "so far no videos were used in the crash investigation."} ... Robin's comments follow claims by two publications, German daily Bild and \green{French Paris Match}, of a cell phone video showing the harrowing final seconds from on board the flight as \blue{it crashed into the French Alps} on March 24 ...
\green{Paris Match and Bild reported that the video was recovered from a phone at the wreckage site. } The two publications described ... 
\\ \hline
\textbf{Summaries: }  \\ \hline
\textbf{[Germanwings Flight 9525]} \blue{Germanwings Flight 9525 crashed into the French Alps.} 
 \\ 
\textbf{[Brice Robin]} \red{Prosecutor Brice Robin says no video footage was used in the crash investigation. 
}
\\
\textbf{[French Paris Match]} \green{French Paris Match says video was found.
}
 \\ \hline
\end{tabular}
\vspace{-0.1in}
\caption{
A sample document and three summaries generated by our entity-controlled model based on DistilGPT2~\cite{DBLP:distilBert2019} and fine-tuned by our proposed method.
Each summary corresponds to the requested entity inside the pair of brackets.
}
\label{figure:intro-example}
\vspace{-0.1in}
\end{figure}

To allow users to control a particular attribute of the generated summaries, \citet{DBLP:conf/aclnmt/FanGA18} proposed a token-based controllable summarization model (ControlSum). Although ControlSum incorporates control tokens that let users specify a requirement on a summary attribute, the maximum likelihood training objective of the model does not provide explicit supervision signals that prevent the model from violating the specified attribute requirement. 
Consequently, a substantial portion of the generated summaries still fail to meet the specified attribute requirement as shown in our experiments. 

One possible solution to enforce the attribute requirement is to apply reinforcement learning (RL) with Markov Decision Process (MDP)~\cite{bellman1957markovian} to optimize a weighted sum of reward functions,
including a penalty function to penalize the violation of the attribute requirement, 
and a summarization metric to encourage the generated summaries to be consistent with the references. 
However, selecting appropriate weights for different reward functions is a delicate task, and requires intensive hyper-parameter tuning.

In this work, we argue that applying constraints on the training objective is a more convenient way to control an attribute of a summary, since it avoids tuning reward function weights. We formulate the problem of training controllable text summarization models as a constrained Markov Decision Process (CMDP)~\cite{altman1999constrained}, a RL framework trained with both rewards and constraints. In this setup, we maximize a summarization metric to encourage the similarity between the output summaries and the references, as well as impose constraints to disallow the summaries from violating a specified attribute requirement.

Moreover, we apply our approach to improve token-based controllable summarization models and control important summary attributes including length, covered entities, and abstractiveness by creating specific constraints for each attribute.
For length control, we divide summary length into disjoint length bins and restrict the summary length according to the desired length bin.
For entity control, we design constraints that guide the generated summary to cover the salient information of user-specified entities.
To control abstractiveness, which measures the degree of textual novelty between a summary and its input document, we define bins corresponding to three abstractiveness levels, and design constraints that allow users to control the summary's abstractiveness.

Extensive experiments are conducted on popular benchmarks, to evaluate the effectiveness of our CMDP training framework with different types of attribute requirements. 
Concretely, we use our CMDP framework to fine-tune controllable summarization models based on pointer-generator network~\cite{DBLP:conf/acl/PtrGen17}, a Recurrent Neural Network (RNN)~\cite{DBLP:journals/neco/HochreiterS97} model, and DistilGPT2~\cite{DBLP:distilBert2019}, a large-scale pre-trained Transformer~\cite{DBLP:conf/nips/VaswaniSPUJGKP17} model\footnote{We choose DistilGPT2 since it is smaller than GPT2. }. 
Experiment results demonstrate that our approach consistently improves both controllable summarization models' capabilities of following the specified attribute requirement.
In addition, our framework increases the ROUGE scores of the generated summaries when provided with the reference control tokens (e.g., the tokens that represent the entities in the reference summary).
Human evaluations further confirm that our framework produces informative summaries that conform to the attribute requirement. 

The key contributions of this paper include: 
(1) A novel training framework that provides explicit guidance signals to supervise a controllable summarization model to conform to the specified attribute requirement; 
(2) Constraints that allow users to control the length, covered entities, and the abstractiveness of the generated summaries, respectively;
(3) Consistent performance improvement of controllable summarization models based on two different architectures.

\section{Related Work}
\noindent \textbf{Summarization systems with specified attributes. }
Several methods extend abstractive summarization models to allow users to control a specific attribute of summaries. 
\citet{DBLP:conf/aclnmt/FanGA18} propose a method that allows users to control an attribute such as length, entity, and style of summaries by prepending special tokens to the input document. 
\citet{DBLP:conf/emnlp/LiuLZ18} focus on controlling the exact length of summaries. They multiply the input word embeddings in the decoder by the specified summary length. 
\citet{DBLP:journals/corr/verbatim-copy19} propose a masked language model to control the portion of copied words in the output summary for the sentence summarization task. This model controls the abstractiveness of a summary at the word level. In contrast, our work controls the extractive fragment density~\cite{DBLP:conf/naacl/newsroom18} of the output summary, which restricts the abstractiveness at the fragment level. 
\citet{DBLP:conf/acl/MakinoITO19} and \citet{DBLP:conf/acl/summaryloop20} incorporate a penalty term on the training objective to penalize a model for violating the length requirement for word limit control. However, it requires hyper-parameter tuning for the weight of penalty if one wants to apply their method to another dataset. Our approach imposes constraints on the training objective and does not need to search suitable weights for penalties based on human inspection. 

Query-focused summarization aims to predict a summary that answers specific questions, e.g., ``\textit{How often did Lebron James visit his hometown?}''. Most of the query-focused summarization methods are extractive and they are based on centrality ranking ~\cite{DBLP:journals/ir/GRSumWan08, DBLP:conf/sigir/CTSumWanZ14}, manifold-ranking ~\cite{DBLP:conf/ijcai/WanYX07, DBLP:conf/ijcai/WanX09, DBLP:conf/cikm/Wan09}, or sentence-compression framework~\cite{DBLP:conf/acl/WangRCFC13}. 
Recently, \citet{DBLP:conf/acl/DiveristyAttn17} propose two query attention-based models for abstractive query-focused summarization. 
On the other hand, entity-controlled summarization aims to produce a summary that captures the salient information of the desired entities, e.g., ``Lebron James''.

\noindent \textbf{Abstractive summarization. }
Most of the existing abstractive summarization models~\cite{DBLP:conf/emnlp/GehrmannDR18, DBLP:conf/icml/pegasus20, DBLP:conf/sigir/ChanReview20} are built on the encoder-decoder model~\cite{DBLP:conf/iclr/BahdanauCB14} to generate summaries. 
\citet{DBLP:conf/acl/PtrGen17} propose the pointer-generator network which allows copying words from the source to the output summary. 
The structure-infused copy mechanism~\cite{DBLP:conf/coling/StructureCopy18} incorporates the syntactic structure of the source text into the pointer-generator network to facilitate copying important words to the output summary. 
\citet{DBLP:conf/acl/Singleton19} propose a summarization framework that first extracts either a single sentence or a pair of sentences from the source document, then it condenses or fuses the selected sentence(s) to generate a summary. The above models do not allow users to constrain the degree of copying nor sentence fusion from the source document. 

Recent methods apply RL with MDP to optimize an abstractive summarization model towards a single or a weighted sum of reward functions. 
Several methods~\cite{DBLP:conf/iclr/PaulusXS18, DBLP:conf/naacl/CelikyilmazBHC18} adopt the ROUGE-L score~\cite{lin2004rouge} as the reward function. 
The SENECA model~\cite{DBLP:conf/emnlp/SharmaHHW19} optimizes a weighted sum of ROUGE-2, ROUGE-L, and a coherence score from a coherence model. 
To improve the factual correctness of the generated summaries, several methods~\cite{DBLP:conf/acl/KGSum20,DBLP:conf/acl/OptimizeFactual20} use RL to maximize a weighted sum of ROUGE scores and a factual correctness score computed by a model. 
\citet{DBLP:conf/emnlp/ImprovingAbstraction18} use the weighted sum of ROUGE-L and $3$-gram novelty as the reward to increase the abstractiveness of summaries, but this method does not allow users to control the abstractiveness level of summaries. 
\citet{DBLP:conf/naacl/SaliencyReward18} extend the ROUGE-L reward by up-weighting the salient words detected by a classifier. One can modify this word-level weighting scheme to encourage the summary to contain certain keywords, but this method does not explicitly encourage the model to generate relevant information about the keywords. 
In contrast, we design a constraint to enforce a summary to retain relevant information of the requested entities. 
\citet{DBLP:human-fine-tune-gpt2} apply RL to fine-tune a GPT2 model~\cite{blog:gpt2}. The reward is provided by a model trained from human preferences on different summaries. 
Though one can use a weighted sum of rewards to control an attribute of generated summaries, such a method needs to tune the weights for rewards. Our CMDP approach avoids the tuning of such weights.

\noindent \textbf{Controllable text generation. }
Controllable text generation has received increasing attention from researchers. 
In machine translation, 
several methods~\cite{DBLP:conf/naacl/SennrichHB16, kobus-etal-2017-domain, DBLP:conf/aclwat/TakenoNY17} apply special tokens to control the politeness, domain, or length of the translation output. 
\citet{DBLP:journals/corr/FiclerG17aa} concatenate a style embedding with the decoder input to control the style of the generated review. 
\citet{DBLP:conf/emnlp/KikuchiNSTO16, DBLP:conf/aaai/MiaoZMYL19, DBLP:conf/acl/SchumannMLVM20} introduce different techniques to control sentence length for the headline generation task, such as feeding a length embedding to the decoder. 
The label-fine-tuning (LFT) model~\cite{DBLP:journals/tacl/NiuB18} uses special tokens to control the politeness of responses for dialogue response generation. 
Several insertion-based decoding methods~\cite{DBLP:conf/cvpr/SunLB17, DBLP:journals/corr/abs-1901-00158, DBLP:journals/tacl/GuLC19} are proposed to complete a fill-in-the-blank sentence, e.g., ``\textit{keywords 1} \_\_ \textit{keywords 2} \_\_''. These decoding methods can be used to enforce the output to contain certain keywords, but users need to specify the relative order among the keywords. In contrast, entity-controlled summarization lets the model determine the relative order among the requested entities. 
Recently, \citet{DBLP:journals/corr/CTRL19} train a large language model conditioned on control codes that specify particular attributes such as domain or language style. Compared with the above methods, our approach incorporates the attribute requirement into the training objective, which gives more explicit supervision signals to the summarizer.

\section{Controllable Summarization with Constrained Markov Decision Process}

\subsection{Problem Definition}
Given a text document $\mathbf{x}$ and a requirement on an attribute $a$ (e.g., length limit of 20 words), the goal of {\it controllable text summarization} is to generate a summary $\mathbf{y}$ that satisfies the requirement. 
Both the input document and output summary are sequences of words, i.e.,  $\mathbf{x}=[x_{1},\ldots,x_{l_{\mathbf{x}}}]$ and $\mathbf{y}=[y_{1},\ldots,y_{l_{\mathbf{y}}}]$, where $l_{\mathbf{x}}$ and $l_{\mathbf{y}}$ are the numbers of words in $\mathbf{x}$ and $\mathbf{y}$ respectively. In this work, we focus on single-document summarization. 

\subsection{Constrained Markov Decision Process Formulation}\label{sec:cmdp}
We propose a constrained Markov Decision Process (CMDP) approach to guide a controllable summarization model to follow the attribute requirement. 
Assume an agent interacts with an environment to generate a summary in discrete time steps. 
At each step $t$, the agent performs an action by sampling a word $y_{t}$ from its policy $\pi_{\bm{\theta}}$, which is a controllable summarization model. 
Then the agent updates its internal state representation (hidden state of the decoder) and proceeds to the next step. 
Once the agent produces the end-of-sequence (EOS) token, we denote the current time step as $T$, the environment gives a reward $r(y_{1},\ldots,y_{T},\mathbf{y}^*,\mathbf{x})$, and a set of costs $c_{i}(y_{1},\ldots,y_{T},\mathbf{y}^*,\mathbf{x})$ to the agent. The process then terminates. 
The reward function $r$ measures the similarity between the output summary $[y_{1},\ldots,y_{T}]$ and the reference summary $\mathbf{y}^*$, while a cost function $c_{i}$ measures how well a summary satisfies an attribute requirement, e.g., we can define a \emph{length cost function} to measure the difference between the output summary length $l_y$ and the specified length limit $l$: $l_y-l$. 
The goal of the agent is to maximize the expected reward while ensuring the costs are under constraints as follows: 
%
\begin{equation}\label{eq:CMDP-obj}
\begin{split}
\underset{\pi_{\bm{\theta}}}{\text{max}} \quad & \mathbb{E}_{\mathbf{y}_{1:T}\sim \pi_{\bm{\theta}}} [r(\mathbf{y}_{1:T},\mathbf{y}^*, \mathbf{x})] \text{,} \\
\text{s.t.} \quad & \mathbb{E}_{\mathbf{y}_{1:T}\sim \pi_{\bm{\theta}}} [c_{i}(\mathbf{y}_{1:T},\mathbf{y}^*,\mathbf{x})] \leq \alpha_{i}, \\ 
  & i=1,\ldots,m \text{,} 
\end{split}
\end{equation}
where $\mathbf{y}_{1:T}$ denotes $y_{1},\ldots, y_{T}$, $\alpha_{i}$ is a pre-defined threshold associated with cost function $c_{i}$, $m$ is the size of the set of constraints. 
A constraint restricts an attribute of the generated summary. For example, to limit the summary length, we can define a constraint to enforce the length cost function to be no larger than 0, $l_y-l\leq 0$. 


\smallskip
\noindent \textbf{Lagrange relaxation. }
Following \citet{DBLP:conf/iclr/RewardConst19}, we apply the Lagrange relaxation technique~\cite{bertsekas1997nonlinear} to approximate the constrained optimization problem in Eq.~(\ref{eq:CMDP-obj}). 
We use $J(\pi_{\bm{\theta}})$ as a shorthand to denote $\mathbb{E}_{\mathbf{y}_{1:T}\sim \pi_{\bm{\theta}}} [r(\mathbf{y}_{1:T}, \mathbf{y}^*, \mathbf{x})]$ and use $J_{c_i}(\pi_{\bm{\theta}})$ to denote $\mathbb{E}_{\mathbf{y}_{1:T}\sim \pi_{\bm{\theta}}} [c_{i}(\mathbf{y}_{1:T}, \mathbf{y}^*, \mathbf{x})]$. 
We then define a Lagrangian function $L(\bm{\lambda}, \bm{\theta})=J(\pi_{\bm{\theta}})- \sum_{i=1}^{m} \lambda_{i} (J_{c_{i}}(\pi_{\bm{\theta}})-\alpha_{i})$, where $\lambda_{i}$ is a Lagrangian multiplier and $\bm{\lambda}=[\lambda_{1},\ldots,\lambda_{m}]\in\mathbb{R}^{m}$. 
When $\lambda_{i}\geq 0$, $\forall i$, the optimal value of $\max_{\bm{\theta}} L(\bm{\lambda}, \bm{\theta})$ is an upper bound to the optimal value of Eq.~(\ref{eq:CMDP-obj}). 
If we minimize the optimal value of $\max_{\bm{\theta}} L(\bm{\lambda}, \bm{\theta})$, we will obtain a tighter upper bound on the optimal value of Eq.~(\ref{eq:CMDP-obj}). Thus, we approximate Eq.~(\ref{eq:CMDP-obj}) by the following relaxed problem: 
%
\begin{align}
\underset{\bm{\lambda} \succeq 0}\min \text{ } & \underset{\bm{\theta}}{\max} & & J(\pi_{\bm{\theta}})- \sum_{i=1}^{m} \lambda_{i} (J_{c_{i}}(\pi_{\bm{\theta}})-\alpha_{i}) \text{,}
\label{eq:relaxed-CMDP}
\end{align}
where $\bm{\lambda} \succeq 0$ denotes that every entry in $\bm{\lambda}$ is non-negative. 
Intuitively, this relaxed problem penalizes the behavior of violating the constraints, and all the Lagrange multipliers $\lambda_{i}$ are learnable. In contrast, the MDP formulation requires the manual tuning of weights for penalty terms. 

\smallskip
\noindent \textbf{Policy training. }
We optimize $\bm{\theta}$ and $\bm{\lambda}$ alternatively using gradient ascent and descent: $\bm{\theta} \leftarrow \bm{\theta} + \eta_{1} \nabla_{\bm{\theta}} L(\bm{\lambda}, \bm{\theta})$, $\bm{\lambda} \leftarrow \bm{\lambda} - \eta_{2} \nabla_{\bm{\lambda}} L(\bm{\lambda}, \bm{\theta})$, where $\eta_{1}$ and $\eta_{2}$ are learning rates for $\bm{\theta}$ and $\bm{\lambda}$ respectively. 
The gradients are expressed as follows. 
$\nabla_{\bm{\theta}} L = \mathbb{E}_{\pi_{\bm{\theta}}} [\sum_{t=0}^{T} \nabla_{\bm{\theta}} \log \pi_{\bm{\theta}}(y_{t}|\mathbf{y}_{1:t-1}) (r - \bm{\lambda}^{T} \mathbf{c})]$, $\nabla_{\bm{\lambda}} L = -(\mathbb{E}_{\pi_{\bm{\theta}}} [\mathbf{c}]- \bm{\alpha})$, where $\mathbf{c}=[c_{1},\ldots,c_{m}] \in \mathbb{R}^{m}$, $\bm{\alpha}=[\alpha_{1},\ldots,\alpha_{m}] \in \mathbb{R}^{m}$, $\mathbb{E}_{\pi_{\bm{\theta}}}$ is a shorthand for $\mathbb{E}_{\mathbf{y}_{1:T} \sim \pi_{\bm{\theta}}}$. 
Since it is intractable to enumerate all possible $\mathbf{y}_{1:T}$, we approximate the expectation $\mathbb{E}_{\mathbf{y}_{1:T} \sim \pi_{\bm{\theta}}}$ using a sample of output sequence $\mathbf{y}_{1:T}\sim \pi_{\bm{\theta}}$. 
Moreover, we also subtract the reward by a baseline $b$, which is a standard technique to reduce the variance of the gradient estimator~\cite{DBLP:books/lib/SuttonB98}. The gradients are then estimated by: 
%
\begin{align}
\nabla_{\bm{\theta}} L &\approx \sum_{t=0}^{T} \nabla_{\bm{\theta}} \log \pi_{\bm{\theta}}(y_{t}|\mathbf{y}_{1:T}) (r - \bm{\lambda}^{T} \mathbf{c} - b) \text{,} \label{eq:theta_gradient} \\
\nabla_{\bm{\lambda}} L &\approx -( \mathbf{c}- \bm{\alpha} ) \text{.} \label{eq:lambda_gradient}
\end{align}
We can interpret $\nabla_{\bm{\theta}} L$ as the standard policy gradient with a regularization term $-\bm{\lambda}^{T} \mathbf{c}$, where $\bm{\lambda}$ is trained by a gradient descent algorithm. 

In this work, we apply the self-critical baseline~\cite{DBLP:conf/cvpr/RennieMMRG17}. Specifically, we use greedy search to generate an output sequence $\bar{\mathbf{y}}$ from the policy. Then, we treat the reward of this sequence $r(\bar{\mathbf{y}}, \mathbf{y}^*, \mathbf{x})$ as the baseline $b$. 

\noindent \textbf{Reward function. }
We apply \textbf{BERTScore}~\cite{DBLP:journals/corr/bert-score} as the reward function to measure the similarity between an output summary and the reference summary based on their BERT~\cite{DBLP:conf/naacl/Bert} contextual embeddings. We do not use ROUGE scores~\cite{lin2004rouge} as the reward since they cannot match paraphrases in an output. 

\noindent \textbf{3-gram repetition constraint. }
Similar to prior work~\cite{DBLP:conf/iclr/PaulusXS18, DBLP:conf/emnlp/LiuL19, DBLP:conf/acl/summaryloop20}, we address the problem of repetition of text fragments by adding a 3-gram repetition constraint into our framework. We define a cost function that measures the ratio of 3-gram repetition in a summary: $\text{RepeatRatio}_{3}(\mathbf{y})=\# repeat\text{ } 3\text{-}gram/\#\text{ }3\text{-}gram$. Then we set its threshold to zero and apply the following 3-gram repetition constraint: $\text{RepeatRatio}_{3}(\mathbf{y}) \leq 0$.

\subsection{Implementation with RNN and Pre-trained Transformer} 
We apply our CMDP framework to train two types of controllable summarization models: pointer-generator network~\cite{DBLP:conf/acl/PtrGen17} and DistilGPT2~\cite{DBLP:distilBert2019}. 
The pointer-generator network is a popular abstractive summarization model based on RNN encoder-decoder model~\cite{DBLP:conf/iclr/BahdanauCB14}. We also incorporate the intra decoder attention~\cite{DBLP:conf/iclr/PaulusXS18} mechanism since it has been shown to improve the performance of the pointer-generator. GPT2~\cite{blog:gpt2} is a large-scale pre-trained language model based on Transformer~\cite{DBLP:conf/nips/VaswaniSPUJGKP17}. DistilGPT2 is a compressed version of GPT2 model using the knowledge distillation technique~\cite{DBLP:distilBert2019}. We append the text ``TL;DR'' to the input document to trigger the summarization operation by DistilGPT2. We append control tokens to these two models. 

\subsection{Length-controlled Summarization}\label{sec:length-control}
Length-controlled summarization aims to control the length of generated summaries. 
We adopt the setting proposed by \citet{DBLP:conf/aclnmt/FanGA18}, which allows users to constrain the summary length to a pre-defined range, e.g., 33 to 37 words. 
We first divide summary length into 10 disjoint length bins $\mathcal{LB}=(lb_{1},\ldots,lb_{10})$. Each length bin corresponds to a range of length, 
and each bin contains a roughly equal number of training samples in the corpus. 
Let $lb_{i^*}$ denote the specified length bin. The goal of this task is to generate a summary $\mathbf{y}$ that satisfies the specified length bin $lb_{i^*}$. 

\noindent \textbf{Base model. }
We expand the vocabulary of the model with ten special tokens (e.g., <bin\_2>) to denote the corresponding bins. In training, we feed the token that indicates the length bin of the reference summary. During testing, we control the length of the output summary by inputting the token of our specified length bin. For pointer-generator, we prepend the token at the beginning of the document. For DistilGPT2, we insert the special token into the ``TL;DR:'' prefix, e.g., ``TL;DR<bin\_2>:''. 

\noindent \textbf{Length bin constraint. }
To encourage the summary length to match the specified length bin, we define a cost function that computes the normalized distance between the length bin of the generated summary $\hat{i}$ and the specified length bin $i^*$: $|\hat{i} - i^*|/10$, then we set the threshold $\alpha=0$, which leads to the following {length bin constraint}: $|\hat{i} - i^*| \leq 0$. We adopt a normalized cost function to prevent the values of costs from being too large and dominating the gradient $\nabla_{\bm{\theta}} L$ in Eq.~(\ref{eq:theta_gradient}). 

\subsection{Entity-controlled Summarization}
\label{sec:entity-control}
Our second task is to generate a summary that focuses on entities requested by a user. 
\citet{DBLP:conf/aclnmt/FanGA18} anonymize each entity in the document by a special token. In contrast, we do not anonymize the entities, which is a more realistic setup. 

\noindent \textbf{Base model. }
During training, we prepend the reference entities to the document. These requested entities are separated by segmenters, e.g., ``Lebron James <ent> LA Lakers''. 
In test time, we control the focus of the summary by feeding in our specified entities. 
To make the reference summaries focus on the reference entities, we remove the reference summary sentences that contain neither reference entities nor coreferent mentions of reference entities on training, validation, and test splits\footnote{Less than 2\% of the removed sentences contain named entities that have coreferent mentions. }. 


\noindent \textbf{QA constraint. }
We apply a question-answering (QA) constraint to guide the generated summary to capture the important information of the requested entities. 
The main idea is to use the QA-based metric from \citet{DBLP:conf/naacl/EyalBE19} and \citet{DBLP:conf/emnlp/ScialomLPS19} to evaluate the capability of a summary to answer a set of questions regarding the reference entities. The QA constraint ensures that the score of the QA-based metric is above a threshold. 

Specifically, we first construct a set of cloze question-answer pairs by individually masking each of the named entities from the reference summary to create the question, with the masked entity as its gold-standard answer. 
The summary predicted by a system is considered as the \textit{context} for a QA model. We feed each of the cloze questions and the context to the QA model, then the QA model extracts an answer from the context for each cloze question. 
We use the $F_{1}$ score of the answers extracted by the QA model as the evaluation metric, denoted as \textbf{QA-}$\bm{F_{1}}$ \textbf{score}. If a summary presents the key information of the reference entities, then the QA-model can predict the correct answers from the summary most of the time. We use the negative of QA-$F_{1}$ as our cost function and set the threshold to -0.9. Our {QA constraint} is then defined as:  $-\text{QA-}F_{1}(\mathbf{y}) \leq -0.9$. 

The QA model is a BERT model~\cite{DBLP:conf/naacl/Bert} with a span classification head on top of the last-layer hidden states. The span classification head is a fully-connected layer that predicts the beginning and ending positions of the answer span on the context. 
We obtain a BERT-based QA model that is fine-tuned on SQuAD 2.0~\cite{DBLP:conf/acl/SQUAD2} from Huggingface Transformers~\cite{Wolf2019HuggingFacesTS}. 
Then we further fine-tune the QA model on the CNN/Dailymail~\cite{DBLP:conf/nips/HermannKGEKSB15, DBLP:conf/conll/NallapatiZSGX16} corpus using our constructed question-context-answer triplets. We construct 349,653/17,442 cloze question-context-answer triplets for training and development. The details of the construction method are described in \S\ref{sec:cloze-construction}.


\noindent \textbf{Entity repetition constraint. }
We find that the QA constraint will cause the model to repeatedly generate the same requested entity in a sentence, 
because the model wants to increase the chance that the QA model will select the requested entities as the answer. Since a named entity usually contains one or two words, the entity repetition behavior cannot be fixed by the 3-gram repetition constraint. 
To address this problem, we first define a function $\text{ER}(\mathbf{y})$ to measure the fraction of sentences in $\mathbf{y}$ that contain repetition of requested entities. We then use $\text{ER}(\mathbf{y})$ as the cost function and apply the following constraint: $\text{ER}(\mathbf{y})\leq 0$.

\subsection{Abstractiveness-controlled Summarization}
Our third task is abstractiveness-controlled summarization, which allows a user to specify the degree of text novelty between a generated summary and the corresponding document\footnote{Abstraction refers to the process of semantic generalization of concepts in the source document. The degree of text novelty is a proxy for measuring abstractiveness.}. 
In this work, we adopt \textbf{extractive fragment density}~\cite{DBLP:conf/naacl/newsroom18} to measure the abstractiveness of a summary. Given a document $\mathbf{x}$ and a summary $\mathbf{y}$, the set of extractive fragments $\mathcal{F}(\mathbf{x}, \mathbf{y})$ is the set of common sequences of words in $\mathbf{x}$ and $\mathbf{y}$. Extractive fragment density is defined as the mean square of the extractive fragment lengths: $\frac{1}{l_{\mathbf{y}}} \sum_{f\in \mathcal{F}(\mathbf{x}, \mathbf{y})} |f|^2$. 
Intuitively, a summary that copies many longer text fragments from the document has a higher extractive fragment density and a lower abstractiveness. 
We divide the values of extractive fragment density into three abstractiveness bins: $ab_{1}=(3.3,+\infty]$, $ab_{2}=(1.3,3.3]$, $ab_{3}=[0,1.3]$, which indicates low, medium, and high abstractiveness respectively. The goal of abstractiveness control is to generate a summary $\mathbf{y}$ that follows the specified abstractiveness bin $ab_{i^*}$. 

\noindent \textbf{Base model. }
Similar to length control, we use special tokens to denote the abstractiveness bins and input a special token to control the abstractiveness level of the output summary. 

\noindent \textbf{Abstractiveness bin constraint. }
To avoid the output summary from violating the specified abstractivenss bin, we apply a cost function to evaluate the normalized distance between the abstractiveness bin of the output summary $\hat{i}$ and the desired abstractiveness bin $i^*$: $|\hat{i} - i^*|/3$. We set the threshold to 0 and obtain the following {abstractiveness bin constraint}: $|\hat{i} - i^*|\leq 0$. 

\noindent \textbf{Conjunction constraint. }
We find that after applying the abstractiveness constraint, the model often inserts the conjunction ``but'' into a copied fragment to decrease the extractive fragment density, even if there is no contrast relationship. 
Since it is difficult to detect the improper use of conjunction, we devise a constraint to avoid the model from generating ``but'' when the reference summary does not contain ``but''. Concretely, we first define a binary function $\text{IC}(\mathbf{y})$ as follows. $\text{IC}(\mathbf{y})=1$ if the predicted summary $\mathbf{y}$ contains ``but'' and the reference summary does not contain ``but''; otherwise, $\text{IC}(\mathbf{y})=0$. We then apply the following conjunction constraint: $\text{IC}(\mathbf{y})\leq 0$. 
This method can be generalized to other discourse markers depending on specific model behavior. 

\section{Experimental Setup}

\noindent \textbf{Datasets. }
We use three popular summarization datasets in our experiments. The first one is the \textbf{CNN/DailyMail}~\cite{DBLP:conf/nips/HermannKGEKSB15,DBLP:conf/conll/NallapatiZSGX16} corpus. We use the standard splits, which have 287,113/13,368/11,490 samples for training, validation, and test sets. Each summary in the training set has 66 words on average. We follow the preprocessing steps of \citet{DBLP:conf/acl/PtrGen17}. Table~\ref{table:ext-frag-density-bin-compare} shows the distribution of abstractiveness bins. We can observe that most of the reference summaries belong to abstractiveness bin 1 and 2, indicating that this dataset is not abstractive. 

Moreover, we use a subset of the \textbf{Newsroom}~\cite{DBLP:conf/naacl/newsroom18} corpus. Newsroom contains 1.3 million news articles with summaries from 38 different news publishers. We construct a subset of the Newsroom corpus called \textbf{Newsroom-b} which has a more balanced distribution of abstractiveness bins. We extract all the samples from three of the news publishers (Washington Post, The Guardian, and New York Times) and obtain the splits of 297,327/31,815/32,047 for training, validation, and test sets. The distribution of abstractiveness bins is shown in Table~\ref{table:ext-frag-density-bin-compare}. 

Furthermore, we conduct experiments of length control on the \textbf{DUC-2002} dataset~\cite{DUC2002} using a test-only setup~\cite{DBLP:conf/emnlp/ChenGTSZY18, DBLP:conf/acl/BansalC18, chan2021condense}. DUC-2002 consists of 567 documents and each document has two reference summaries. We remove the documents that are shorter than their corresponding reference summaries, resulting in 554 documents. This dataset has long reference summaries with an average length of 113 words. 

\begin{table}[t]
\centering
\small
\fontsize{10}{11}\selectfont
\begin{tabular}{c|ccc}
\hline \hline
Bin       & CNN/DM & Newsroom & Newsroom-b \\
\hline \hline
1             & 37.88\%       & 45.92\%  &  33.94\% \\
2             & 57.56\%       & 25.96\%  &  37.54\% \\
3             & 4.56\%        & 28.12\%  &  28.52\% \\
\hline
\end{tabular}
\caption{
Distribution of abstractiveness bins of reference summaries on CNN/DM, Newsroom, and Newsroom-b training sets. Bin 3 is the most abstractive bin. Newsroom-b is a subset of Newsroom which has a more \emph{balanced} distribution of abstractiveness bins. 
}
\label{table:ext-frag-density-bin-compare}
\vspace{-0.15in}
\end{table}

\noindent \textbf{Baselines and comparison. }
We use maximum likelihood (ML) loss to train the pointer-generator and DistilGPT2 based controllable summarization models described in \S\ref{sec:entity-control}, denoted as \textbf{PG} and \textbf{D.GPT2} respectively. 
We then use a suffix ``+CMDP'' to indicate that a model is fine-tuned by our CMDP framework. 
The following baselines do not use pre-trained models. 
We consider the \textbf{ControlSum}~\cite{DBLP:conf/aclnmt/FanGA18} model as a baseline for all of our control settings. 
For entity control, we incorporate query-focused summarization baselines including \textbf{GRSUM}~\cite{DBLP:journals/ir/GRSumWan08}, an extractive model that incorporates query-relevance into a random walk algorithm, \textbf{QueryAtt}~\cite{DBLP:conf/acl/DiveristyAttn17}, an abstractive model that applies a query attention to focus on different parts of the input query, and \textbf{SD2}~\cite{DBLP:conf/acl/DiveristyAttn17}, which integrates an orthogonality constraint into the QueryAtt model to encourage the successive query attention context vectors to be orthogonal to each other. Both the QueryAtt and SD2 models have a strong inductive bias that the generated summary should focus on the query. 
We modify the ROUGESal~\cite{DBLP:conf/naacl/SaliencyReward18} method by doubling the weights to the words of the requested entities and treat it as a baseline, denoted as \textbf{ROUGEEnt}. 


\begin{table*}[t]
\centering
\small
\fontsize{9}{10}\selectfont
\resizebox{\textwidth}{!}{
\begin{tabular}{l|cccccccccccc}
\hline \hline
       & \multicolumn{3}{c}{Bin 1} & \multicolumn{3}{c}{Bin 4} & \multicolumn{3}{c}{Bin 7} & \multicolumn{3}{c}{Bin 10} \\
Method      & R-1    & R-2     & R-L    & R-1     & R-2      & R-L    & R-1     & R-2     & R-L    & R-1     & R-2   & R-L    \\
\hline \hline
ControlSum        & 32.40    & 14.30    & 28.28    & 36.30    &    15.34  &  31.95    & 38.55   & 16.18  &  34.50   & 40.30   & 17.08 &  36.59  \\ \hline
PG     & 27.93     &  12.06     &  24.40    & 31.41        &  12.51        &  27.23    & 31.81        &   12.27      &   27.54    & 31.94       & 11.79      &  28.09       \\
PG+CMDP    & \textbf{35.30}   & \textbf{17.00}     & \textbf{31.98}     & {37.88}      &    \textbf{17.59}      &    {34.27}    & \textbf{39.85}    &  \textbf{18.46}      & \textbf{36.17}      &    40.73 &  17.11    &    37.30    \\ \hline
D.GPT2    & 31.21   & 13.36     & 27.12     & 36.27      &    15.97      &    31.91    &  38.18   &  16.43      & 33.64      &    40.87 & {17.45}      &    36.62    \\
D.GPT2+CMDP    & 33.09   & 13.48    & 29.74    &  \textbf{38.41}     &     16.55     &  \textbf{34.59}   & {39.65}    &  16.77      & {35.79}      & \textbf{42.05}    & \textbf{17.77}      &    \textbf{38.35}    \\ \hline
\end{tabular}
}
\caption{
Results of length control on different specified length bins using the DUC-2002 data. 
Our CMDP framework consistently improves the ROUGE scores of PG and D.GPT2 ($p<0.04$, approximate randomization test, for ROUGE-1 and ROUGE-L). 
}
\label{table:len-bin-DUC-results}
\vspace{-0.1in}
\end{table*}

\noindent \textbf{Evaluation metrics. }
For length control and entity control, we evaluate the quality of summaries using \textbf{ROUGE-1}, \textbf{ROUGE-2}, and \textbf{ROUGE-L} $F_{1}$ scores with full-length and stemming~\cite{lin2004rouge}. 
For abstractiveness control, we use embedding-based metrics, \textbf{BERTScore}~\cite{DBLP:journals/corr/bert-score} and \textbf{MoverScore}~\cite{DBLP:conf/emnlp/ZhaoPLGME19}, to measure the semantic similarity between an output summary and a reference summary. To evaluate how well the generated summaries satisfy the attribute requirement, we define a metric called \textbf{bin \%} to measure the percentage of generated summaries that follow the specified bin (length or abstractiveness bin). We use the \textbf{QA-}$\bm{F_{1}}$ score defined in \S\ref{sec:entity-control} to evaluate whether a summary retains the essential information of the reference entities. 
We define \textbf{reference entities} as all the named entities (typed as location, person, and organization) that appear in both the reference summary and the first 400 words of the input document. 
We also define \textbf{appear \%} to measure the percentage of requested entities that appear in the summary. 
For the non-reference control settings, the entire test set is evaluated under different control constraints and reference summaries do not exist in these cases. 

\noindent \textbf{Implementation Details. }
We use Spacy~\cite{spacy} for coreference resolution. 
For RNN-based models, we use the Adam algorithm~\cite{DBLP:journals/corr/ADAM14} for training. We first use ML loss to train a RNN-based model until the validation loss stops decreasing for three consecutive checkpoints. Then we start the (C)MDP training. 
The initial learning rates are 1e-3 and 5e-5 for ML and CMDP training respectively. 
For Transformer-based models, we use the AdamW algorithm~\cite{DBLP:journals/corr/ADAMW} for training. We first use ML loss to train a Transformer-based model for 12 epochs. Then we start the (C)MDP training. The initial learning rates are 5e-5 and 1.77e-5 for ML and CMDP training. During CMDP training of D.GPT2, we freeze the bottom four layers of the model. We initialize the values of $\bm{\lambda}$ to 0.01. 

\section{Automatic Evaluation Results}

\subsection{Results of Length Control}
\noindent \textbf{Reference length bin. }
We first evaluate the performance of length controlled models when supplying the length bin of the reference summary (reference length bin) at testing time. 
The results are shown in Table~\ref{table:len-bin-cnn-results}. 
We observe that \textit{after applying our CMDP framework, both PG and D.GPT2 models obtain significantly higher ROUGE scores and a larger portion of their generated summaries follow the specified length bin}. We also report the results of the D.GPT2 model after fine-tuned by RL with MDP (D.GPT2+MDP). In this MDP approach, the reward is BERTScore minus a weighted sum of length bin distance and 3-gram repetition ratio. We tune the weights of penalties on the validation set and set the weights for length bin distance and 3-gram repetition to 0.4 and 0.6 respectively. We can see that our CMDP approach outperforms the MDP approach. The above results demonstrate the effectiveness of our framework. 

\begin{table}[t]
\centering
\small
\fontsize{9}{10}\selectfont
\begin{tabular}{l|cccc}
\hline \hline
Method                       & R-1   & R-2   & R-L   &  Bin \% \\
\hline \hline
ControlSum                  & 39.75     & 17.43     & 36.70     & 48.15 \\ \hline
PG                        & 35.07     & 15.05     & 32.11     & 74.09 \\
PG+CMDP                   & 39.77     & 16.65     & 36.66     & \textbf{94.37} \\ \hline
D.GPT2                        & 39.28     & 17.36     & 36.07     & 50.74 \\
D.GPT2+CMDP                   & \textbf{41.72}     & \textbf{17.99}     & \textbf{39.00}     & 70.13 \\ \hline
D.GPT2+MDP                   & 41.46     & 17.69     & 38.74     & 69.71 \\
\hline
\end{tabular}
\caption{
Results of length control using reference length bins as the input on the CNN/DM dataset. 
Our CMDP framework significantly improves the ROUGE scores and bin \% of both PG and D.GPT2 ($p<0.0001$, approx. randomization test).
}
\label{table:len-bin-cnn-results}
\vspace{-0.1in}
\end{table}

Moreover, we observe that the D.GPT2 based models obtain higher ROUGE scores but lower bin \% than the PG based models. One possible reason is that the large-scale pre-training in D.GPT2 makes the model more difficult to adapt to a specific bin requirement. This suggests a trade-off between the task metrics and the bin \%. 

\noindent \textbf{Arbitrary length bin. }
We evaluate the performance of length-controlled models when supplying different length bins at testing time. We report the results of length-controlled models on four different length bins: 1, 4, 7, and 10. The DUC-2002 dataset is adopted since this dataset has long reference summaries. Hence, we can evaluate the quality of summaries with different lengths by truncating the summaries. We truncate the reference and system summaries to 33, 46, 59, and 100 for specified length bins of 1, 4, 7, and 10 respectively when computing ROUGE scores. 
ROUGE evaluation with truncation is a common practice for evaluating a system summary when given a length budget~\cite{DBLP:conf/lrec/HongCFKLN14}. The intuition is that a good summary should contain the more essential information at the beginning. 

\begin{figure}[t]
\centering
\includegraphics[width=\linewidth]{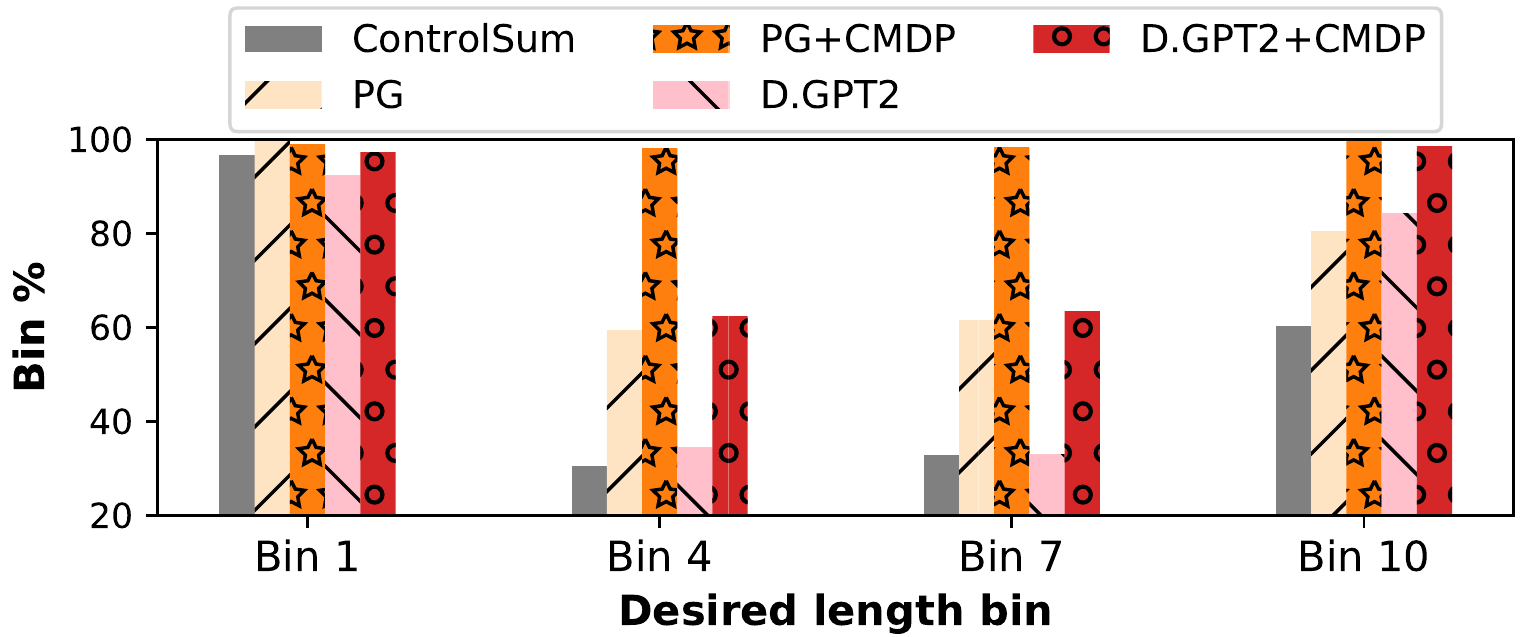}
\vspace{-0.2in}
\caption{
Bin \% of different models with different specified length bins on the DUC-2002 dataset. 
Our framework improves the bin \% of PG and D.GPT2 for bin 4, 7, and 10 by a wide margin. 
}
\label{fig:len-bin-percent-compare}
\vspace{-0.1in}
\end{figure}

We analyze the results of length-controlled models on different length bins. Figure~\ref{fig:len-bin-percent-compare} illustrates the results of bin \% obtained by different models. We observe that \textit{all the models achieve more than 90 bin \% for length bin 1}. It is because length bin 1 represents the range of $(0,33]$ in length, it is easy to satisfy the requirement by generating a very short summary. 
\textit{For length bin 4, 7, and 10, our CMDP framework improves the bin \% of both PG and D.GPT2 models by a wide margin}. 
From Table~\ref{table:len-bin-DUC-results}, we can see that \textit{our framework consistently improves the ROUGE scores of PG and D.GPT2 models}. 

\begin{figure}[t]
\centering
\includegraphics[width=\linewidth]{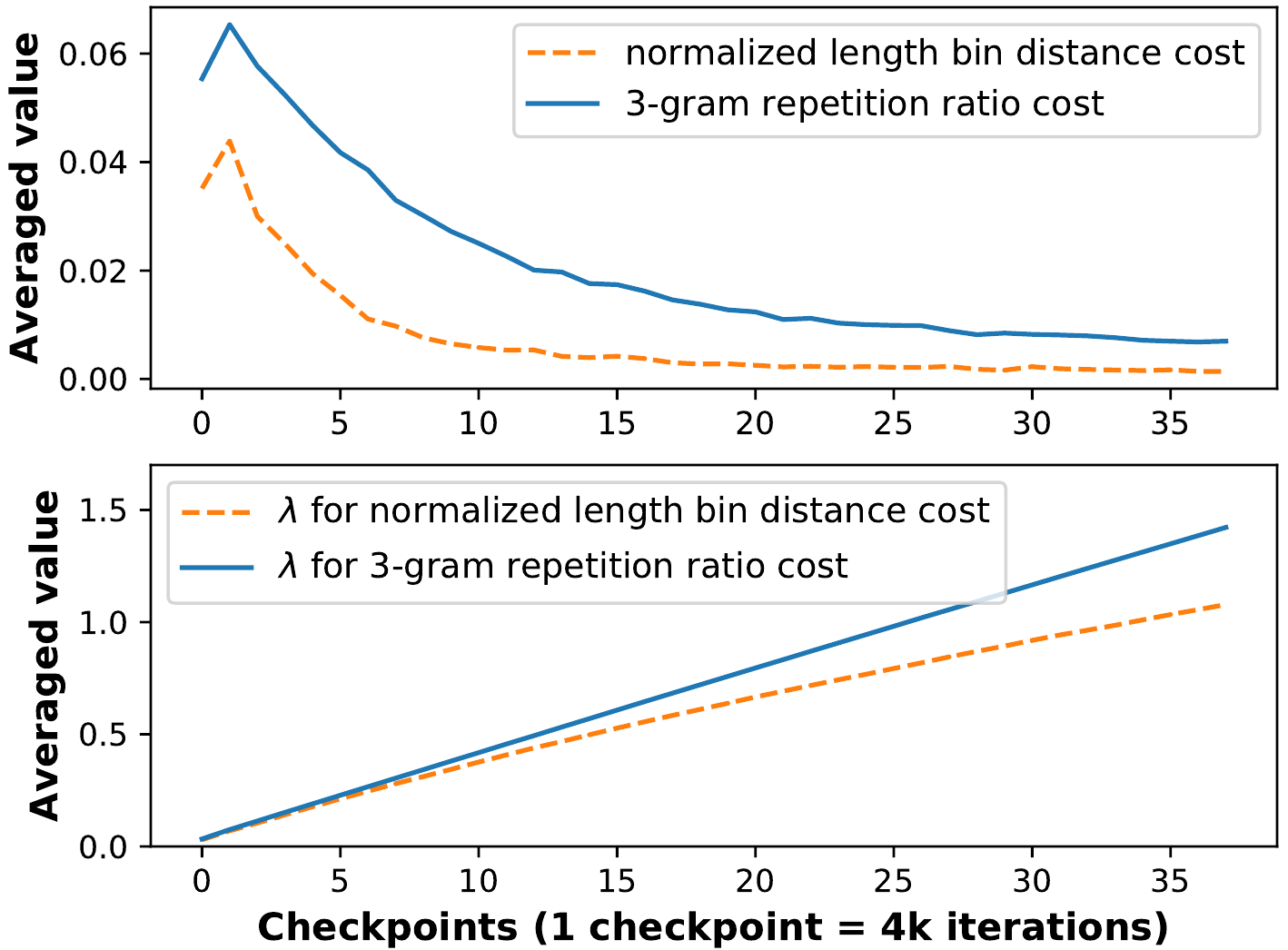}
\vspace{-0.2in}
\caption{
Values of costs ($\mathbf{c}$) and Lagrangian multipliers ($\bm{\lambda}$) of PG+CMDP for length control on every checkpoint (4k iterations) during training. Each value is averaged over 4k iterations.
}
\label{fig:len-bin-cost-lambda}
\vspace{-0.1in}
\end{figure}

\noindent \textbf{Costs and Lagrangian multipliers. }
Furthermore, we analyze the values of costs ($\mathbf{c}$) and Lagrangian multipliers ($\bm{\lambda}$) of our PG+CMDP model during training. From Figure~\ref{fig:len-bin-cost-lambda}, we can see that \textit{the costs received by the agent decrease gradually over iterations}. It is because the relaxed training objective of our framework in Eq.~(\ref{eq:relaxed-CMDP}) penalizes the behavior of violating the constraints. 
We also observe that \textit{the values of Lagrangian multipliers $\bm{\lambda}$ keeps increasing}. 
The reason is that according to Eq.~(\ref{eq:lambda_gradient}), the gradient of $\bm{\lambda}$ is negative as long as there is a sample that violates the constraints during training. As mentioned in \S~\ref{sec:cmdp}, $\bm{\lambda}$ is learned by a gradient descent algorithm and the algorithm increases $\bm{\lambda}$ when the gradient is negative. 

\begin{table}[t]
\centering
\small
\fontsize{9}{10}\selectfont
\resizebox{\columnwidth}{!}{
\begin{tabular}{l|cccccc}
\hline \hline
Method                       & R-1   & R-2   & R-L & QA-$F_1$  & Appear \% \\
\hline \hline
GRSUM                      & 35.89    & 15.86  & 31.96 &  34.92  &  76.22 \\ 
ROUGEEnt                   & 39.45    & 20.36  & 36.78 &  23.47  &  83.75 \\ 
ControlSum                 & 39.41    & 19.94  & 36.55 &  27.02  &  74.08 \\ 
QueryAtt                  & 38.92     & 20.38  & 36.47 &  25.12  &  75.10 \\ 
SD2                       & 39.43     & 20.71  & 36.88 &  27.23  &  75.97 \\ \hline
PG                        & 37.61     & 19.27  & 35.04 &  23.53  &  37.96 \\ 
PG+CMDP                   & 40.81     & 20.23  & 37.56 &  30.38  & 86.64 \\  \hline
D.GPT2                     & 41.68     & 22.32  & 38.85 &  35.32  &  82.31 \\
D.GPT2+CMDP                & \textbf{45.00}  &  \textbf{23.65}  &  {41.85} &  \textbf{36.00} & {93.37} \\ \hline
D.GPT2+MDP                & \textbf{45.00}  &  23.50  &  \textbf{41.90} &  35.72 & \textbf{94.46} \\
\hline
\end{tabular}
}
\caption{
Results of entity-controlled models using reference entities as the input on the CNN/DM dataset. 
Our CMDP framework significantly improves the ROUGE scores, QA-$F_1$, and appear \% ($p<0.0001$, approx. randomization test). 
}
\label{table:gold-entity-results}
\vspace{-0.1in}
\end{table}

\subsection{Results of Entity Control}\label{sec:exp-auto-entities}
\noindent \textbf{Reference entities. }
We first evaluate the performance of entity-controlled models in summarizing the reference entities. 
For each of the models, we feed in all the reference entities to generate a summary that centers on the reference entities. 
The results are presented in Table~\ref{table:gold-entity-results}. We use the CNN/DM dataset for entity-controlled summarization because it contains named entities in 99.74\% of the reference summaries in its test set, whereas the Newsroom-b dataset only has 85.24\%. 
When computing QA-$F_{1}$ and appear \%, we ignore the samples that do not have a named entity in the reference summary. We observe that \textit{our framework consistently and significantly improves the ROUGE scores, QA-$F_{1}$ score, and appear \% for both of the PG and D.GPT2 models}. 
These results demonstrate the effectiveness of our framework in summarizing reference entities. 

\begin{table*}[t]
\centering
\small
\fontsize{9}{10}\selectfont
\begin{tabular}{l|ccccccccc}
\hline \hline
       & \multicolumn{3}{c}{bin 1} & \multicolumn{3}{c}{bin 2} & \multicolumn{3}{c}{bin 3} \\
Method      & BERTS.  & MoverS. & Bin \%  & BERTS.   & MoverS. & Bin \%  & BERTS. & MoverS. &  Bin \%    \\
\hline \hline
ControlSum  & 26.53    & 16.15 & 99.40   & 24.50 & 13.11     & 6.42   & 20.53  & 9.87  & 24.30 \\
\hline
PG          & 26.49    & 15.99 & 99.36   & 22.59 & 11.15     & 8.49   & 17.67  & 7.45  & 26.03 \\
PG+CMDP     & 29.62    & 18.44  & \textbf{99.72}   & 26.95 & 12.96  & \textbf{97.65}   & 22.78  & 7.67  & \textbf{98.88}  \\
\hline
D.GPT2        & 27.85 & 17.05    & 99.17  & 26.41    & 14.57  &  0.47 &  22.30   &  11.16  & 00.37   \\
D.GPT2+CMDP   & \textbf{30.12} &  \textbf{18.45}   & {99.72}  &  \textbf{31.21}   & \textbf{17.52}   & 72.48 &  \textbf{25.75} &  \textbf{13.18}  &  87.09 \\
\hline
D.GPT2+MDP   & {29.77} &  {18.27}   & \textbf{99.77}  &  {30.40}  & {17.06}   &  72.37 &  {25.25} &  {13.06}  &  80.95 \\
\hline
\end{tabular}
\caption{
Results of abstractiveness-controlled models with different specified bins on Newsroom-b dataset. 
Bin 3 is the most abstractive bin. 
Our CMDP framework significantly improves the BERTScore, MoverScore, and bin \% over all the bins ($p<0.003$, approx. randomization test). 
}
\label{table:ext-frag-newsroom-results}
\vspace{-0.1in}
\end{table*}

We also adopt the D.GPT2+MDP model as a rival system. In this control setting, the reward is $\text{BERTScore}(\mathbf{y})+\gamma_1\text{QA}F_1(\mathbf{y}) -\gamma_2 \text{RepeatRatio}_{3}(\mathbf{y}) -\gamma_3 \text{ER}(\mathbf{y})$. 
We set $\gamma_1,\gamma_2,\gamma_3$ to 0.15, 0.4, and 0.5 respectively after hyper-parameter tuning. It is observed that the MDP approach and our CMDP approach obtain similar performance while our approach has fewer hyper-parameters to tune. 

\begin{figure}[t]
\centering
\includegraphics[width=\linewidth]{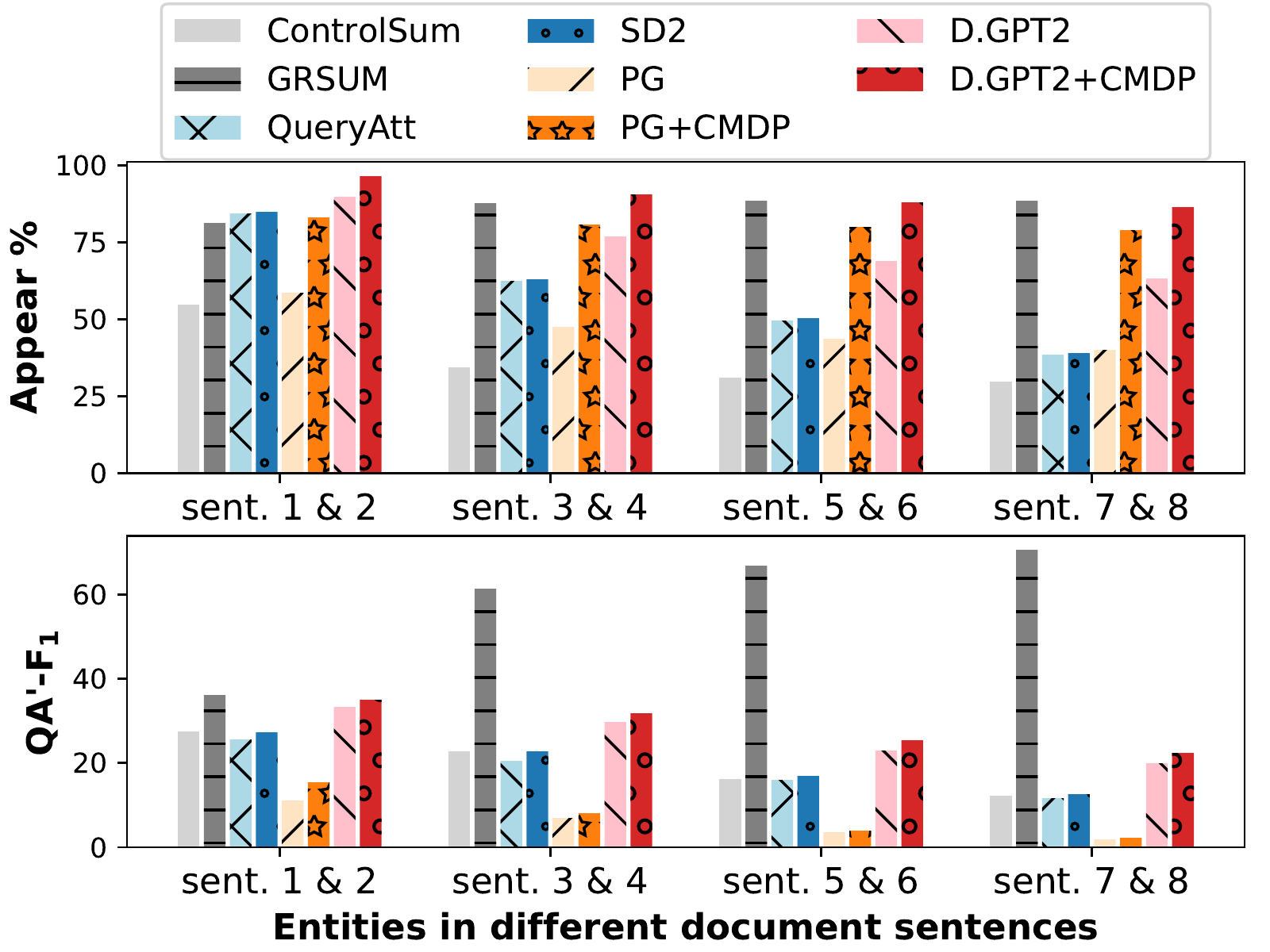}
\caption{Results of entity-controlled models for entities in different document sentences. 
Our CMDP framework consistently improves the QA$'$-$F_1$ and appear \% for entities at different positions. 
}
\label{fig:entities-by-positions}
\vspace{-0.1in}
\end{figure}

\noindent \textbf{Entities at different positions. }
Next, we evaluate the capability of entity-controlled models to summarize entities at different positions of the document with the following setup. 
For each of these models, we use the named entities at document sentences 1 to 2, 3 to 4, 5 to 6, and 7 to 8 as the requested entities respectively. 
Since we do not have reference summaries for these entities, we use the document sentences to construct cloze questions to evaluate the output summaries. For each requested entity, we build cloze questions by masking each document sentence that contains the entity or its coreferent mention. We use the $F_{1}$ score of the answer predicted by the QA model as an evaluation metric, denoted as QA$'$-$F_{1}$.

We analyze the performance of our method for entities at various sentences of the document. 
The results of appear \% and QA$'$-$F_{1}$ scores are presented in Figure~\ref{fig:entities-by-positions}. 
We observe that \textit{our CMDP framework consistently improves the appear \% and QA$'$-$F_{1}$ scores of both PG and D.GPT2 models for entities at different positions}. 
Without our CMDP training, the appear \% are low for entities at latter positions of the document. 
The reason is that we use reference entities for model training and the reference entities are concentrated in the first few sentences of the document, which bias a neural model towards these sentences. 
There are 45.6\% of reference entities appear in the first two document sentences in the training set of CNN/DM. 
Nevertheless, \textit{the neural models fine-tuned by our CMDP achieve high appear \% for entities at varying positions}. 

Moreover, we observe that the GRSUM system achieves highest QA$'$-$F_{1}$ scores and its appear \% scores are similar to that of D.GPT2+CMDP. We analyze the reasons as follows. The GRSUM system is an extractive method while all other methods in Figure 4 are abstractive methods. It is relatively easy for an extractive method to select document sentences that mention the request entities to obtain high appear \%. In the setting of non-reference entity control, we use document sentences to construct the cloze questions for the QA$'$-$F_{1}$ metric since we do not have a reference summary. Hence, the QA$'$-$F_{1}$ metric tends to give higher scores to extractive summaries. Moreover, we also observe that the GRSUM model achieves higher QA$'$-$F_{1}$ scores for the entities at latter sentences of the document. 
The entities at latter positions of a news article are usually less important entities that are only mentioned once and do not have coreferent mentions. The GRSUM system relies on term vectors to measure the relevance of a sentence. Thus, this system cannot recognize a coreferent mention that uses completely different words (e.g., pronoun). As a result, it is easier for GRSUM to extract a summary for entities at latter locations. 
However, an extractive method cannot paraphrase the information of the document to generate a concise entity-focused summary.

\begin{table*}[t]
\centering
\small
\fontsize{9}{10}\selectfont
\begin{tabular}{l|ccccccccc}
\hline \hline
       & \multicolumn{3}{c}{bin 1} & \multicolumn{3}{c}{bin 2} & \multicolumn{3}{c}{bin 3} \\
Method      & BERTS. & MoverS. & Bin \%  & BERTS.   & MoverS. & Bin \%  & BERTS. & MoverS. &  Bin \%    \\
\hline \hline
ControlSum  & 38.55    & 23.56 & 99.94   & 39.47 & 23.23     & 1.09   & 37.51  & 20.58  & 0.03 \\
\hline
PG          & 35.42    &  20.00 &  99.85   & 34.91    & 18.67  & 1.42   & 32.53  & 15.87  & 0.21 \\
PG+CMDP     & 41.77    & 25.96  &  \textbf{100.00}  & 40.79    &  23.54    &  \textbf{75.10}  &  34.22  &  17.71  &  \textbf{48.62}  \\
\hline
D.GPT2        & 39.02 &  23.82   & 99.90  &  39.58   &  23.30 & 1.93 & {38.15}  &  21.23  &  0.01   \\
D.GPT2+CMDP   & \textbf{43.23} &  \textbf{26.65}   & 99.56  &  \textbf{44.07}   &  \textbf{26.39}  & {62.60} &  \textbf{42.03} &  \textbf{24.71}  &  {1.94} \\
\hline
D.GPT2+MDP   & {42.56} &  {26.43}   & 99.77  &  {43.59}   &  {26.23}  & {55.67} &  {41.44} &  {24.42}  &  {2.09} \\
\hline
\end{tabular}
\caption{
Results of abstractiveness-controlled models with different specified bins on CNN/DM dataset. Our CMDP framework significantly improves the BERTScore and MoverScore ($p<0.003$, approx. randomization test) over all the bins. 
It also significantly improves the bin \% for bin 2 and 3 ($p<0.00001$, approx. randomization test). 
}
\label{table:ext-frag-CNN-results}
\vspace{-0.1in}
\end{table*}

\subsection{Results of Abstractiveness Control}\label{sec:abs-results-auto}
We analyze the capability of abstractiveness-controlled models to generate summaries with different abstractiveness levels. 
In our experiments, for each of the abstractiveness-controlled models, we feed in abstractiveness bin 1, bin 2, and bin 3 independently. 
The results on Newsroom-b and CNN/DM datasets are presented in Table~\ref{table:ext-frag-newsroom-results} and \ref{table:ext-frag-CNN-results}. 
We can see that \textit{our CMDP framework consistently improves the BERTScores and MoverScores of PG and D.GPT2 models}. 
We also observe that \textit{all the models achieve more than 99 bin \% for bin 1 (least abstractive)}, because it is easier for models to directly copy document sentences than to paraphrase document information. \textit{For abstractiveness bin 2 and 3, our CMDP framework substantially improves the bin \% of PG and D.GPT2 models}, which show that our framework improves the ability of summarization models to generate summaries of higher abstractiveness levels. 
Similar to the results of length control, there is a trade-off between the task metrics and the bin \%. 

We then compare the bin \% results on the CNN/DM dataset with that on Newsroom-b. It is observed that for abstractiveness bin 3 (most abstractive), all the models achieve a low bin \% on CNN/DM but a substantially higher bin \% on Newsroom-b. This is because in the CNN/DM, there are only 4.6\% of the training samples belonging to bin 3. Hence, it is difficult for a model to learn to generate a highly abstractive summary. 
In contrast, the Newsroom-b dataset has a balanced distribution of abstractiveness bins so that a model can learn from more abstractive references. 

\begin{table}[t]
\centering
\small
\fontsize{10}{11}\selectfont
\resizebox{\columnwidth}{!}{
\begin{tabular}{l|ccc}
\hline \hline
Method              &  Fluency  & Entity-rel.  & Faithful.  \\
\hline \hline
SD2                 & 4.83     & 3.63   &  70\%  \\ 
D.GPT2              & 4.65     & 3.33   &  68\%  \\ 
D.GPT2+CMDP         & 4.83     & 3.92   &  71\%  \\ \hline
\end{tabular}
}
\caption{
Human fluency, entity-relevance, and faithfulness scores of entity-controlled models with the reference entities as the input. Faithful. denotes the percentage of generated summaries that are faithful. 
The Krippendorf's $\alpha$ inter-rater agreement for all columns are 0.68, 0.77, and 0.56. 
}
\label{table:ent-control-compare-human}
\vspace{-0.1in}
\end{table}

Furthermore, we compare our framework with the D.GPT2+MDP model on both datasets. 
The reward is $\text{BERTScore}(\mathbf{y})-\gamma_1 |\hat{i}-i^*|/3 -\gamma_2 \text{RepeatRatio}_{3}(\mathbf{y}) -\gamma_3 \text{IC}(\mathbf{y})$, where $\hat{i}$ denotes the abstractiveness bin of the generated summary and $i^*$ denotes the specified abstractiveness bin. On the CNN/DM dataset, we set $\gamma_1,\gamma_2,\gamma_3$ to 0.3, 0.5, and 0.3 respectively. On the Newsroom-b dataset, we set these weights to 0.4, 0.5, and 0.3 respectively. We observe that the MDP approach and our CMDP approach obtain similar performance while our approach has fewer hyper-parameters to tune. 

\section{Human Evaluation}
We conduct human evaluation to verify the quality of the generated summaries. 
We hire postgraduate students as annotators and each test sample is evaluated by three annotators. The names of models are blinded to the annotators. 

\subsection{Results of Entity Control}
The human annotators evaluate entity-controlled summarization models using the following metrics: (i) \textbf{fluency}: estimating the readability and grammaticality of a summary using a rating from 1 to 5; (ii) \textbf{faithfulness}: a yes/no question indicating whether a summary is factually consistent with the document. The annotators are instructed to state ``yes'' only if the summary does not contain any factual inconsistencies; and (iii) \textbf{entity-relevance}: evaluating how well a summary retains the key information of the requested entities from 1 to 5.

\noindent \textbf{Reference entities. }
We ask human annotators to evaluate the quality of summaries when requesting reference entities. 
For each of the entity-controlled models, we feed in all the reference entities. 
The overall number of annotators is six. 
For each of the test samples, we present the input document, requested entities, reference summary, and three system summaries generated by SD2, D.GPT2, and D.GPT2+CMDP models. 
We present the evaluation scores on 100 random samples of the CNN/DM dataset in Table~\ref{table:ent-control-compare-human}. For the faithfulness metric, we report the percentage of faithful summary computed by majority vote (i.e., at least two out of three annotators vote as faithful). 
Our D.GPT2+CMDP method significantly outperforms the D.GPT2 and SD2 models in terms of entity-relevance (power analysis with mixed effects model~\cite{DBLP:conf/emnlp/PowerAnalysis20}, power $>0.99$, approx. randomization test, $p<0.0001$) while maintaining similar fluency and faithfulness with the SD2 model (approx. randomization test, $p>0.97$). 

\noindent \textbf{Entities at different positions. }
We pick the best two models (SD2 and D.GPT2+CMDP) in the previous section to further conduct human evaluation for entities at different sentences of the document. 
The total number of annotators is four. 
As mentioned in \S\ref{sec:exp-auto-entities}, most of the reference entities are located in document sentences 1 to 2. To avoid too much overlapping with the reference entities setting, we do not choose the bin of sentences 1 to 2 and conduct evaluation on the subsequent two bins, sentences 3 to 4 and 5 to 6. 
For each model, we feed in the named entities at document sentences 3 to 4 and 5 to 6 as the requested entities respectively. 
Since we do not have gold-standard summaries for this setup, we cannot show the reference summaries to the annotators. 
The results on 100 random samples are shown in Table~\ref{table:non-ref-ent-compare-human}. 
Our D.GPT2+CMDP model consistently achieves higher entity-relevance scores than the SD2 model (power analysis with mixed effects model, power $>0.81$, approx. randomization test, $p<0.0001$) and obtains competitive fluency and faithfulness scores (approx. randomization test, $p>0.41$). 

\begin{table}[t]
\centering
\small
\fontsize{10}{11}\selectfont
\resizebox{\columnwidth}{!}{
\begin{tabular}{l|l|ccc}
\hline \hline
Sent. & Method              &  Fluen.  & Ent.-rel.  & Faith.  \\
\hline \hline
\multirow{2}{*}{3\&4} & SD2                 & 4.75     & 2.81   &  63\%  \\ 
& D.GPT2+CMDP         & 4.79     & 3.36   &  64\%  \\ \hline
\multirow{2}{*}{5\&6} & SD2                 & 4.78     & 2.68   &  62\%  \\ 
& D.GPT2+CMDP         & 4.78     & 3.29   &  62\%  \\ \hline
\end{tabular}
}
\caption{
Human fluency, entity-relevance, and faithfulness scores of entity-controlled models for entities at different document sentences. 
The Krippendorf's $\alpha$ inter-rater agreement for these scores are 0.60, 0.78, and 0.44. 
}
\label{table:non-ref-ent-compare-human}
\vspace{-0.1in}
\end{table}

\subsection{Results of Abstractiveness Control}
The annotators evaluate abstractiveness-controlled models using the following setting. 
There are six annotators for the results of CNN/DM dataset and three annotators for the results of Newsroom-b dataset. For each test sample, we generate two groups of system summaries (group 1 and group 2). 
For group 1, we use our D.GPT2+CMDP model to generate three different summaries by feeding abstractiveness bin 1, bin 2, and bin 3 respectively. For group 2, we use our PG+CMDP model to generate three different summaries using a similar method. 
During evaluation, we present the source document, the reference summary, and two groups of system summaries to the annotators. The summaries within each group are randomly shuffled. 

\begin{table}[t]
\centering
\small
\fontsize{10}{11}\selectfont
\begin{tabular}{l|cccc}
\hline \hline
       & \multicolumn{2}{c}{CNN/DM} & \multicolumn{2}{c}{Newsroom-b} \\
Method              & EM  & PM  &  EM  & PM  \\
\hline \hline
PG+CMDP             & 66\%  & 94\%   &  84\%    &  96\%   \\ 
D.GPT2+CMDP         & 66\%  & 92\%   &  86\%    &  98\%   \\ \hline
\hline
\end{tabular}
\caption{
Results of exact match (EM) and partial match (PM) scores of human abstractiveness rankings that are consistent with the specified bins. 
The Krippendorf's $\alpha$ inter-rater agreement for the abstractiveness rankings on CNN/DM and Newsroom-b are 0.85 and 0.72 respectively. 
}
\label{table:abs-control-rank}
\vspace{-0.1in}
\end{table}

\noindent \textbf{Abstractiveness among summaries. }
We evaluate the abstractiveness of the generated summaries by human judgments using the following setup. For each group of system summaries, we ask the annotators to give a ranking among the three system summaries according to their abstractiveness. For instance, if an annotator thinks that summary 1 > summary 2 > summary 3 in terms of abstractiveness, then the annotator gives a ranking of $[3, 2, 1]$ to them. 
The abstractiveness rankings from different annotators are then aggregated by averaging. 
If the aggregated abstractiveness ranking is consistent with the order of our specified abstractiveness bins, then this group of summaries has an \textbf{exact match}. For example, suppose the order of our specified abstractiveness bins is $[3, 2, 1]$. If the aggregated abstractiveness ranking is $[3, 1.6, 1.3]$, then then this group of summaries has an exact match. If the aggregated abstractiveness ranking is $[3, 1.3, 1.6]$, then there is no exact match. 
Moreover, we investigate whether the summaries of abstractiveness bin 1 and bin 3 can be distinguished by annotators. If the aggregated abstractiveness ranking is consistent with the order of abstractiveness bin 1 and bin 3, then there is a \textbf{partial match}. 
Suppose the order of our specified abstractiveness bins is $[3, 2, 1]$, if the aggregated ranking is $[3, 1.3, 1.6]$, then there is a partial match. If the aggregated ranking is $[1.6, 1.3, 3]$, then there is no partial match.

We analyze the exact match and partial match scores of abstractiveness-controlled models as follows. 
The results on 100 random test samples of the CNN/DM and Newsroom-b datasets\footnote{We use both CNN/DM and Newsroom-b because we want to understand the impact of the training dataset on the abstractiveness of the output summaries.} are presented in Table~\ref{table:abs-control-rank}. We observe that our models on both of the two datasets achieve very high partial match scores, but our models on the CNN/DM dataset obtain lower exact match scores than that on the Newsroom-b dataset (approx. randomization test, $p<0.02$). 
This is because the CNN/DM dataset is extractive in nature. Hence, it is more difficult to learn three levels of abstractiveness on CNN/DM. Nonetheless, our models can still achieve more than 60\% exact match scores.

\begin{table}[t]
\centering
\small
\fontsize{10}{11}\selectfont
\resizebox{\columnwidth}{!}{
\begin{tabular}{l|l|ccc}
\hline \hline 
Bin                   & Method      & Flu. & Rel. & Faithful. \\ \hline \hline 
\multirow{2}{*}{1} & PG+CMDP    & 4.79    & 3.43    &   98\%         \\
                   & D.GPT2+CMDP & 4.75    &  3.34  &   96\%        \\ \hline 
\multirow{2}{*}{2} & PG+CMDP    & 4.52   & 2.34     &   58\%        \\
                   & D.GPT2+CMDP & 4.57   & 3.14    &   66\%          \\ \hline 
\multirow{2}{*}{3} & PG+CMDP    & 4.47   & 2.00     &   52\%         \\
                   & D.GPT2+CMDP & 4.60   & 2.99    &   66\%       \\ \hline 
\end{tabular}
}
\caption{
Human fluency, relevance, and faithfulness scores of abstractiveness-controlled models on Newsroom-b. 
The Krippendorf's $\alpha$ inter-rater agreement for these metrics are 0.51, 0.37, and 0.40. 
}
\label{table:abs-control-individual}
\vspace{-0.1in}
\end{table}

\noindent \textbf{Quality of individual summaries. }
Next, we ask the annotators to evaluate the qualities of the summaries of three different abstractiveness bins using the following metrics: (i) \textbf{fluency}: measuring the readability of a summary from 1 to 5; (ii) \textbf{faithfulness}: a yes/no question asking whether a summary is factually consistent with the document; and (iii) \textbf{relevance}: evaluating how well a summary retains the salient information of the document on 1-5. 
The results of 100 random test samples from the Newsroom-b dataset\footnote{We choose Newsroom-b because there are more generated summaries that satisfy the abstractiveness bin requirement, which is more suitable for comparing the quality of summaries of different abstractiveness bins. } are presented in Table~\ref{table:abs-control-individual}. When using abstractiveness bin 1 (lowest level), all the models achieve significantly higher fluency, relevance, and faithfulness (approx. randomization test, $p<0.005$). 
The scores of all these metrics drop substantially for abstractiveness bin 2 and bin 3 because paraphrasing is more challenging than copying. 
Figure~\ref{figure:newsroom-abs-case-study} illustrates sample summaries generated by our D.GPT2+CMDP model on the Newsroom-b dataset. 
We observe that the generated summary of bin 3 has a factual error, which is italicized in the figure.

\begin{figure}[t]
\small
\centering
\fontsize{9}{10}\selectfont
\begin{tabular}{|p{0.92\columnwidth}|}
\hline
\textbf{Reference Summary:} \\ 
\blue{Manchester United have announced a 10-year contract with German manufacturers Adidas to be the club's new kit sponsor for a record-breaking minimum \pounds750m.}
\\ 
\hline
\textbf{Abstractiveness bin 1 (least abstractive):} \\ 
\blue{Manchester United have announced a 10-year contract with the German manufacturers Adidas to be the club's new kit sponsor for a record-breaking minimum \pounds750m.}
\\ \hline
\textbf{Abstractiveness bin 2 (medium abstractive):} \\ 
\blue{Manchester United have announced a 10-year contract with} Adidas to sponsor \blue{new kit sponsor} for \pounds750m club record signing.
\\ \hline
\textbf{Abstractiveness bin 3 (most abstractive):} \\ 
\blue{Manchester United} \red{are hoping to secure} \blue{\pounds750m} sponsorship deal with German company after contract is signed in \blue{2015-16 season}.
\\ \hline
\end{tabular}
\caption{
Sample summaries generated by our D.GPT2+CMDP model with abstractiveness bin 1, 2, and 3 on the Newsroom-b testing set. Extractive fragments in summaries are in blue color. Factual errors are in red color. 
}
\label{figure:newsroom-abs-case-study}
\vspace{-0.15in}
\end{figure}

\section{Conclusion}
We propose a novel CMDP training framework for controllable text summarization. Our framework imposes constraints on the training objective to explicitly disallow the output summaries from violating the requirement specified by users. Moreover, we apply our framework to control key summarization attributes such as length, covered entities, and abstractiveness of the summaries. We then devise specific constraints to restrict each of these attributes respectively. Empirical studies on popular benchmarks demonstrate that our framework significantly improves the capability of controllable summarization models to conform to the desired attribute requirement. 

In our framework, we can set hard constraints without tuning threshold values. For instance, we set the threshold of our length bin constraint to 0 to disallow the violation of length bin requirement. 
Compared to the weights of penalty in the MDP framework, the threshold value in a soft constraint is also easier to set. For example, the goal of entity control is to generate a summary that presents the key information of the requested entities, which implies that the generated summaries should obtain a high QA-F1 score. The range of QA-F1 score is [0, 1]. In order to encourage the generated summaries to obtain a high QA-F1 score, the threshold for QA-F1 score should be close to 1, this gives us a clue about how to set the value of threshold. On the other hand, the MDP framework does not give us any clues to set the values of penalty weights. In summary, our CMDP framework needs to tune one threshold value for entity control and it does not need to tune any threshold for other control settings. Whereas the numbers of penalty weights to be tuned in the MDP framework are 2, 3, and 3 for length, entity, and abstractiveness control respectively.

\appendix
\section{Appendix}
\subsection{Output Samples for Length Control}
Figure~\ref{figure:cnn-len-case-study} presents sample summaries generated by our D.GPT2+CMDP model using different length bins on the DUC-2002 testing set. We observe that our model discards secondary information when given a shorter length budget. 

\begin{figure}[t]
\small
\centering
\fontsize{10}{11}\selectfont
\begin{tabular}{|p{0.92\columnwidth}|}
\hline
\textbf{Length bin 0:} \\ 
Hurricane Gilbert slams into Kingston on Monday.
115 mph winds cause flash floods and mud slides.
Jamaica expected to receive 10 inches of rain.
Hurricane warnings are canceled.
\\ \hline
\textbf{Length bin 3:} \\ 
NEW: Hurricane Gilbert slams into Kingston on Monday with 115 mph winds.
No serious injuries were reported in the city of 750,000.
The hurricane lashed Kingston's airport and aircraft.
Jamaica will receive 10 inches of rain.
Hurricane warnings are canceled.
\\ \hline
\textbf{Length bin 6:} \\ 
NEW: Hurricane Gilbert slams into Kingston on Monday with 115 mph winds.
No serious injuries were reported in the city of 750,000 people.
The hurricane hit Kingston's airport and aircraft parked on its fields.
The National Weather Service reports heavy damage to Kingston's airport.
Jamaica will receive 10 inches of rain.
\\ \hline
\textbf{Length bin 9:} \\ 
NEW: ``The eye is going to move lengthwise across that island,'' a man says.
Hurricane Gilbert slams into Kingston on Monday with 115 mph winds.
The National Weather Service reports heavy damage to Kingston's airport and aircraft parked on its fields.
``People were running around in the main lobby of our hotel,'' a man says.
Jamaica is expected to receive 10 inches of rain.
Hurricane warnings are canceled in Cuba. 
\\ \hline
\end{tabular}
\caption{
Sample summaries generated by our D.GPT2+CMDP model using different length bins on the DUC-2002 testing set. 
}
\label{figure:cnn-len-case-study}
\vspace{-0.15in}
\end{figure}


\subsection{Training data for the QA Model}\label{sec:cloze-construction}
We construct question-context-answer triplets to train a QA model. 
We individually mask each named entity in a reference summary to create a cloze question and the masked entity is its answer. The reference summary is used as the context. For example, suppose the reference summary $y^*$ is ``Arsenal beat Chelsea 3-1 yesterday.'', then we construct two cloze questions, $q_1=$``[MASK] beat Chelsea 3-1 yesterday.'' and $q_2=$``Arsenal beat [MASK] 3-1 yesterday.'', and two answers, $a_1=$``Arsenal'' and $a_2=$``Chelsea''. After that, we obtain two question-context-answer triplets, $(q_1, y^*, a_1)$ and $(q_2, y^*, a_2)$. 

Since the constructed cloze questions are too similar to the corresponding reference summaries, if we only use reference summaries as the context in our training data, it will encourage the QA model to only rely on surface clues to extract answers. 
To alleviate this problem, we use the method by \citet{DBLP:conf/acl/BansalC18} to extract a pseudo reference summary $\tilde{y}$ from the source document. Then we use $\tilde{y}$ as the context to construct another set of question-context-answer triplets $\{(q_i,\tilde{y},a_i)\}$. The pseudo reference summary includes the document sentences that achieve highest ROUGE-L recall with the reference summary. We discard a triplet if $\tilde{y}$ does not contain all the named entities in the reference. 
To have a balanced training data, we only keep the training triplets $(q_i,{y}^*,a_i)$ that has a corresponding pseudo reference summary $(q_i,\tilde{y},a_i)$. 

To allow the QA model to give a prediction of ``unanswerable'' to low-quality summaries, we construct two types of unanswerable training samples: irrelevant training samples and repeated-entity training samples. 
For irrelevant training samples, we select document sentences that do not contain the reference entities and have a low textual overlap with the reference summary (ROUGE-L recall $\leq 0.2$). 
For repeated-entity training samples, we find out the sentences in the reference summary that contains two named entities and repeat one of its named entities. We treat such samples as unanswerable since they contain factual inconsistencies. 
Overall, our training data consists of 109,815 unanswerable samples and 239,838 answerable samples. 
We will release our training data for the QA model.

\section*{Acknowledgements}
The work described in this paper was partially supported by the Research Grants Council of the Hong Kong Special Administrative Region, China (CUHK 2410021, Research Impact Fund, R5034-18), National Key Research and Development Program of China (No. 2018AAA0100204), the Science and Technology Development Fund of Macau SAR (File no. 0015/2019/AKP), and Guangdong-Hong Kong-Macao Joint Laboratory of Human-Machine Intelligence-Synergy Systems (No. 2019B121205007). 
Lu Wang is supported in part by the National Science Foundation through a CAREER award IIS-2046016. 
We would like to thank the action editor and the anonymous reviewers for their comments. 

\bibliography{tacl2018}
\bibliographystyle{acl_natbib}

\end{document}